%% file: main.tex
\documentclass{article}

\usepackage{arxiv}

\usepackage[utf8]{inputenc} 
\usepackage[T1]{fontenc}    
\usepackage{hyperref}       
\usepackage{url}            
\usepackage{booktabs}       
\usepackage{amsfonts}       
\usepackage{nicefrac}       
\usepackage{microtype}      
\usepackage{xcolor}         
\usepackage{natbib}
\usepackage{times}
\usepackage{epsfig}
\usepackage{graphicx}
\usepackage{amsmath}
\usepackage{amssymb}
\usepackage{bbm}
\usepackage{microtype}
\usepackage{comment}
\usepackage{tabularray}

\usepackage{comment}

\usepackage{algorithm}
\usepackage{algpseudocode}
\usepackage{mathtools}

\def\z{{\mathbf{z}}}

\def\s{{\mathbf{s}}}
\def\a{{\mathbf{a}}}
\def\g{{\mathbf{g}}}
\def\c{{\mathbf{c}}}

\title{A Minimalist Prompt for Zero-Shot Policy Learning}

\author{
 Meng Song \\
 UC San Diego 
 \And
 Xuezhi Wang\\
 Google Deepmind
 \And
 Tanay Biradar\\
 UC Santa Barbara
 \And
 Yao Qin\\
 UC Santa Barbara
 \And
 Manmohan Chandraker \\
 UC San Deigo\\
}
\date{}

\begin{document}

\maketitle

\begin{abstract}
Transformer-based methods have exhibited significant  generalization ability when prompted with target-domain demonstrations or example solutions during inference. Although demonstrations, as a way of task specification, can capture rich information that may be hard to specify by language, it remains unclear what information is extracted from the demonstrations to help generalization. Moreover, assuming access to demonstrations of an unseen task is impractical or unreasonable in many real-world scenarios, especially in robotics applications. These questions motivate us to explore what the minimally sufficient prompt could be to elicit the same level of generalization ability as the demonstrations. We study this problem in the contextural RL setting which allows for quantitative measurement of generalization and is commonly adopted by meta-RL and multi-task RL benchmarks. In this setting, the training and test Markov Decision Processes (MDPs) only differ in certain properties, which we refer to as task parameters. We show that conditioning a decision transformer on these task parameters alone can enable zero-shot generalization on par with or better than its demonstration-conditioned counterpart. This suggests that task parameters are essential for the generalization and DT models are trying to recover it from the demonstration prompt. To extract the remaining generalizable information from the supervision, we introduce an additional learnable prompt which is demonstrated to further boost zero-shot generalization across a range of robotic control, manipulation, and navigation benchmark tasks.      
\end{abstract}

\section{Introduction}
Learning skills that are generalizable to unseen tasks is one of the fundamental problems in policy learning and is also a necessary capability for robots to be widely applicable in real-world scenarios. Recent works \cite{saycan, llm-zero-shot, rt2, reasoning} have shown that transformer based policies can be easily steered to solve new tasks by conditioning on task specifications (prompts). Tasks can be specified in various and complementary forms such as language instructions \cite{saycan, generalist-agent}, goal images \cite{zero-shot-passive-video, zero-shot-visual-il} and demonstrations \cite{prompt-dt, hyper-dt, one-shot-visual-il}. Among typical task specifications, prompting with demonstration is one of the most expressive interfaces to communicate task objectives and human intent to the agent. Even a segment of demonstration contains rich information about environment dynamics, success states, and optimal behaviors to solve the task when expert-level demonstrations are provided. Demonstrations can also be represented in the original action and state space, which does not require any modality alignment. In addition, demonstrations are more favorable in describing goal locations and behaviors of low-level control than language instructions. However, specifying tasks via demonstrations suffers from two major limitations.   

Firstly, it is impractical or unreasonable to have access to any form of demonstrations in test environments, or even refer to expert demonstrations of solving the test task beforehand. In some sense, directly providing the agent with demonstrations during deployment is akin to revealing the solutions to the student in a test, thereby making the generalization meaningless. An appropriate task specification should be zero-shot, that is, conveying what to do instead of how to do it. For example, in a kitchen scenario, one may ask a robot to move a bottle onto the table. A natural task specification is to simply provide the robot with a description of the task rather than hinting at the possible trajectory of its arm and the target pose of its gripper when the task succeeds. This motivates us to distill the task description which is key to the generalization from a demonstration prompt and remove the skill-related information. Secondly, although a demonstration prompt encompasses rich information, there is little understanding of which factors are essential for the generalization. For example, given a visual demonstration of a running half-cheetah, it is unclear which factor contributes to the generalization -- the appearance of the cheetah, the dynamics, the gait, or the desired speed. 

To answer these questions, we consider the contextual RL setting where the training and test MDPs are assumed to be drawn from the same distribution which is controlled by a set of task parameters. These parameters can describe any properties of an MDP, which could either be physics-level attributes such as texture, location, velocity, mass, and friction, or semantic-level attributes, such as object color, shape, quantity, and category. These quantified task variations provide the necessary information for the agent to generalize across different tasks. Therefore, prompting an agent with these task parameters serves as an interpretable baseline for generalization. We show that conditioning a decision transformer on these task parameters alone generalizes on par with or better than its
demonstration-conditioned counterpart. This suggests that a DT model implicitly extracts task parameters from the demonstration prompt to allow generalization to a new task. 

Besides investigating the necessary role of task parameters in how the generalization happens, we are also interested in how much generic information can be elicited from the supervision to help generalization. To this end, we additionally introduce a learnable prompt to further enhance the generalization performance. The learnable prompt and the task parameter together compose a minimalist prompt for zero-shot generalization. We have empirically shown that this proposed prompt has strong zero-shot generalization capability across a variety of control, manipulation, and navigation tasks.

In summary, the main contribution of this paper is a study of what is a minimalist prompt for decision transformers to learn generalizable policies on robotics tasks. 
(1) We show that a task parameter prompting decision transformer can generalize on par with or better than its state-of-the-art few-shot counterpart -- the demonstration prompting DT in a zero-shot manner. This observation identifies the task parameter as necessary information for generalization rather than the demonstrations. (2) We propose to use an additional learnable prompt to extract the generic information from the supervision and show that this information can further boost the generalization performance.

\section{Related Work}
{\bf Generalization in policy learning}
Learning a policy that can be generalized to novel tasks has been extensively explored under different learning objectives and conditions. For example, meta-reinforcement learning \cite{pearl} trains an RL agent to rapidly adapt to new tasks where the agent is enabled to continue to learn from its online interactions with the test environments. \cite{darla} trains a source RL policy operating in a disentangled latent space and directly transfers it to the visually different target domain. Recent works \cite{macaw} extend meta-RL to an offline setting where the agent is fine-tuned on pre-collected transitions in test tasks for adaptation. Unlike these prior works focusing on optimizing the RL objective, we are interested in understanding the generalization behavior of a policy learned under an imitation learning objective, i.e. behavior cloning. 

{\bf Few-shot generalization}
Generalization in the imitation learning setting can be categorized into two categories: few-shot and zero-shot generalization. Few-shot methods condition the model on demonstrations or example solutions from the test environments to enable generalization. For example, \cite{prompt-dt, hyper-dt, generalist-agent} conditions their model on the trajectory segments from expert demonstrations of the test tasks. \cite{zero-shot-visual-il} solves the test task by matching the visual observations along the demonstration trajectories. \cite{one-shot-il, one-shot-visual-il} requires a single demonstration to generalize to new tasks although the demonstration could be performed by an agent with different morphology. \cite{zero-shot-passive-video} transfers the skills learned from human videos to a new environment by conditioning on the goal image of the successful state. Learning a few-shot policy is akin to in-context learning \cite{rethink_demonstration}, but differs in that it is not an inference-only method.

{\bf Zero-shot generalization}
In contrast, zero-shot methods assume no access to reference solutions in the test environments which is more realistic. Possible ways to specify a task in a zero-shot way include natural language instructions \cite{bc-z, saycan, alfred}, driving commands \cite{zero-shot-driving}, programs \cite{program_mani}, etc. In this work, we demonstrate that task parameters, such as the target position of an object and the goal velocity, can sufficiently specify a task to enable zero-shot generalization. 

{\bf Goal-conditioned policy}
Learning a goal-conditioned policy can also be viewed as a way of generalization. The \emph{goal} here usually refers to a goal state indicating the success of a task. A goal-conditioned policy is hoped to reach any goal state including those unseen during training and can be solved as online RL \cite{uvfa, her, gcsl}, offline RL \cite{offline-rl}, and imitation learning problems \cite{goal-state-gen-long-horizon, goal-condition-il}. Mathematically, we can draw a connection between goal-conditioned policy and prompting a language model. We can think of the goal state as a special prompt when solving a goal-conditioned imitation learning problem with the policy instantiated as a transformer-based model. 

{\bf Decision transformer} Decision transformer \cite{dt} (DT) is one of the successful sequence modeling methods \cite{big-sequence-model} in solving RL via supervised learning \cite{rvs} problems. It combines the advantages of return-conditioned policy learning \cite{upsidedown-rl, reward-condition-policy, rvs} and the expressive capability of GPT2. DT has been extended to solve offline multi-task learning \cite{multi-dt, generalized-dt}, online RL \cite{online-dt} problems. Complementary to the prior works of using DT in few-shot policy learning \cite{prompt-dt, hyper-dt}, our method enables DT-based policy to generalize zero-shot to new tasks.

{\bf Adapting language models in the robotics domain} 
Recent work has shown that pre-trained large language models (LLMs) can be used in the robotics domain to ground skills and enable emergent reasoning via supplying high-level semantic knowledge to the downstream tasks \citep{saycan, palme, rt2}. In a similar spirit, our work can be viewed as training a language model agent from scratch to perform robotics tasks with essential task parameters and a learnable prompt as instructions. 

{\bf Prompt learning} A series of work \citep{li-liang-2021-prefix, prompt-tuning} in NLP has shown that a pre-trained language model can be adapted to downstream tasks efficiently via learning prefixes or soft prompts, which is called \textit{prompt tuning}. In this paper, we adapt this technique to learn a generalizable policy which we refer to as \textit{prompt learning}. We condition the decision transformer on a learnable prompt to extract the generic information for zero-shot generalization. Note that in our case, the prompt is learned jointly with the language model during training and both are frozen during inference, whereas in prompt tuning, the soft prompt is only trained during inference while the language model is frozen.

\section{Problem Formulation} 
{\bf Contextual RL framework} We study the problem of policy generalization under the framework of contextual RL \cite{contextual_rl}, where each task $T$ corresponds to a MDP denoted by a tuple $\mathcal{M}=\left(\mathcal{S}, \mathcal{A}, \mathcal{P}, \rho_{0}, r, \gamma\right)$. $\mathcal{S}$ denotes the state space, $\mathcal{A}$ denotes the action space, $\mathcal{P}\left(\mathbf{s}_{t+1} \mid \mathbf{s}_{t}, \mathbf{a}_{t}\right)$ is a conditional probability distribution defining the dynamics of the environment. $\rho_0(\mathbf{s}_0)$ is the initial state distribution, $r: \mathcal{S} \times \mathcal{A} \rightarrow \mathbb{R}$ defines a reward function, and $\gamma \in (0, 1]$ is a scalar discount factor.

In contextual RL setting, a policy is trained on $N$ training tasks $\{T_i \}_{i=1}^N$ and tested on $M$ unseen tasks $\{T_j \}_{j=1}^M$. All training and test tasks are assumed to be drawn from the same distribution over a collection of MDPs. Each MDP $\mathcal{M}(\c)$ in the collection is controlled by a task parameter $\c$, which can parameterize either the dynamics $\mathcal{P}\left(\mathbf{s}_{t+1} \mid \mathbf{s}_{t}, \mathbf{a}_{t}\right)$, the reward function $r(\mathbf{s}_t, \mathbf{a}_t)$, the state space $\mathcal{S}$, and action space $\mathcal{A}$.

Contextual RL framework is expressive enough to formulate many generalization problems in policy learning. A typical example of such problems is to learn a goal-reaching policy \cite{her, goal-condition-il} that can reach any goal state $\g \in \mathcal{S}$. This can be formulated as training a goal-conditioned policy under an indicator reward function of having the current state exactly match the goal state. In this case, the task parameter $\c$ is the goal state $\g$. The definition of transfer learning in the contextual RL setting can also cover the cases where the generalization happens across different dynamics \cite{non-stationary}, state spaces \cite{zero-shot-passive-video}, and action spaces \cite{cycle-consist, morph_transfer}.

{\bf Behavior cloning (BC) in contextual RL setting} In this paper, we consider the problem of learning a generalizable policy under the behavior cloning objective in a contextual RL setting. In this case, each training task $T_i$ is paired with a dataset of demonstration trajectories $\mathcal{D}_i^{\text{train}} = \{\tau_k\}_{k=1}^K (i=1,\cdots, N)$. The entire training set $\mathcal{D}^{\text{train}}=\{\mathcal{D}_i^{\text{train}}\}_{i=1}^N$ were pre-collected in each training environment and could be of any quality. 

The objective of BC is to estimate a policy distribution $\hat{\pi}$ by performing supervised learning over $\mathcal{D}^{\text{train}}=\{(\s_{i}^{\star}, \a_{i}^{\star})\}_{i=1}^n$:
\begin{equation} \label{eq:bc}
    \widehat{\pi}=\operatorname{argmin}_{\pi \in \Pi}\mathcal{L}(\pi) =\operatorname{argmin}_{\pi \in \Pi} \sum_{i=1}^{n} \mathcal{L} (\s_{i}^{\star}, \a_{i}^{\star};\pi)
\end{equation}
where we set the loss function $\mathcal{L}$ as squared $\ell_2$ distance $\mathcal{L} (\s_{i}^{\star}, \a_{i}^{\star};\pi) = \| \pi(\s_{i}^{\star}) - \a_{i}^{\star} \|_2^2$.   

{\bf Evaluation metrics}
For a fair comparison, we follow the conventions in \cite{d4rl, generalist-agent} of evaluating the policy performance across different tasks. The performance is reported as a normalized score:
\begin{equation} \label{def:score}
    \text { normalized score }=100 \times \frac{\text { return-random return}}{\text { expert return-random return}}
\end{equation}
where $100$ corresponds to the average return of per-task expert and $0$ to the average return of a random policy.  

The performance of a trained agent can be evaluated in two settings: (1) {\bf Seen tasks}: The agent is evaluated in each of $N$ training tasks, which is consistent with the standard RL and IL evaluation protocols. (2) {\bf Unseen tasks}: The agent is evaluated in each of $M$ test tasks which are \emph{strictly unseen} during training to measure its zero-shot generalization ability.

\section{Approach}
\subsection{Decision transformer}
Recently, there has been an increasing interest in casting policy learning problems as supervised sequential modeling problems \cite{rvs, big-sequence-model}. By leveraging the power of transformer architecture and hindsight return conditioning, Decision Transformer-based models \cite{dt, multi-dt} have achieved great performance in offline policy learning problems.

Formally, Decision Transformer (DT) is a sequence-to-sequence model built on a transformer decoder mirroring GPT2 \cite{gpt} architecture, which applies a causal attention mask to enforce autoregressive generation. DT treats an episode trajectory $\tau$ as a sequence of 3 types of input tokens: returns-to-go, state, and action, i.e. $\tau =\left(\widehat{R}_{0}, \s_{0}, \a_{0}, \widehat{R}_{1}, \s_{1}, \a_{1}, \ldots, \widehat{R}_{T-1}, \s_{T-1}, \a_{T-1}\right)$, where the returns-to-go $\widehat{R}_{t}=\sum_{i=t}^{T} r_{i}$ computes the cumulative future rewards from the current step until the end of the episode. At each time step $t$, it takes in a trajectory segment of $K$ time steps $\tau_t = \left(\widehat{R}_{t-K+1}, \s_{t-K+1}, \a_{t-K+1}, \ldots, \widehat{R}_{t-1}, \s_{t-1}, \a_{t-1}, \widehat{R}_{t}, \s_{t}, \a_{t}\right)$ and predicts a $K$-step action sequence $A_t = \left(\hat{\a}_{t-K+1}, \ldots, \hat{\a}_{t-1}, \hat{\a}_{t}\right)$.

A DT parameterized by $\theta$ instantiates a policy $\pi_{\theta}(\a_t | \mathbf{s}_{-K,t}, \mathbf{a}_{-(K-1),t-1},\widehat{R}_{-K,t})$, where $\mathbf{s}_{-K, t}$ denotes the sequence of $K$ past states $\mathbf{s}_{\max(0, t-K+1): t}$. Similarly,  $\widehat{R}_{-K, t}$ denotes the sequence of $K$ past returns-to-go, and $\mathbf{a}_{-(K-1),t-1}$ denotes the sequence of $K-1$ past actions. By conditioning on a sequence of decreasing returns-to-go $\mathbf{\widehat{R}}_{-K, t}$, a DT policy becomes aware of its distance to the target state under a measure defined by the reward function.

Given an offline dataset of n K-length trajectory segments $\mathcal{D}=\{\tau^*_i\}_{i=1}^n$. The policy is trained to optimize the following behavior cloning objective:
\begin{equation} \label{eq:dt_bc}
 \operatorname{argmin}_{\theta}\mathcal{L}(\theta) =\operatorname{argmin}_{\theta} \sum_{i=1}^{n} \mathcal{L} (\tau_{i}^{\star};\pi_\theta)
\end{equation}

Specifically, we use the following squared $\ell_2$ distance as the loss function in the continuous domain:
\begin{equation} \label{eq:dt_l2}
\mathcal{L} (\tau_{i}^{\star};\pi_\theta) = \sum_{j=1}^K \|\pi_{\theta}(\a_j | \mathbf{s}^{i*}_{-K,j}, \mathbf{a}^{i*}_{-(K-1),j-1},\widehat{R}^{i*}_{-K,j}) - \a^{i*}_j\|^2_2
\end{equation}

During the evaluation, a target returns-to-go $\widehat{R}_{0}$ should be specified at first. This represents the highest performance we expect the agent to achieve in the current task.  At the beginning of an evaluation episode, DT is fed with $\widehat{R}_{0}$, $\s_0$ and generates $\a_0$, which is then executed in the environment and $\s_1$, $r_1$ are observed by the agent. We then compute $\widehat{R}_{1}=\widehat{R}_{0}-r_1$ and feed it as well as $\s_1$, $\a_1$ to the DT. This process is repeated until the episode ends.

\subsection{Minimalist Prompting Decision Transformer}
\begin{figure}[t!]
    \begin{minipage}[c]{\columnwidth} 
        \centering 
        \includegraphics[width=\columnwidth]{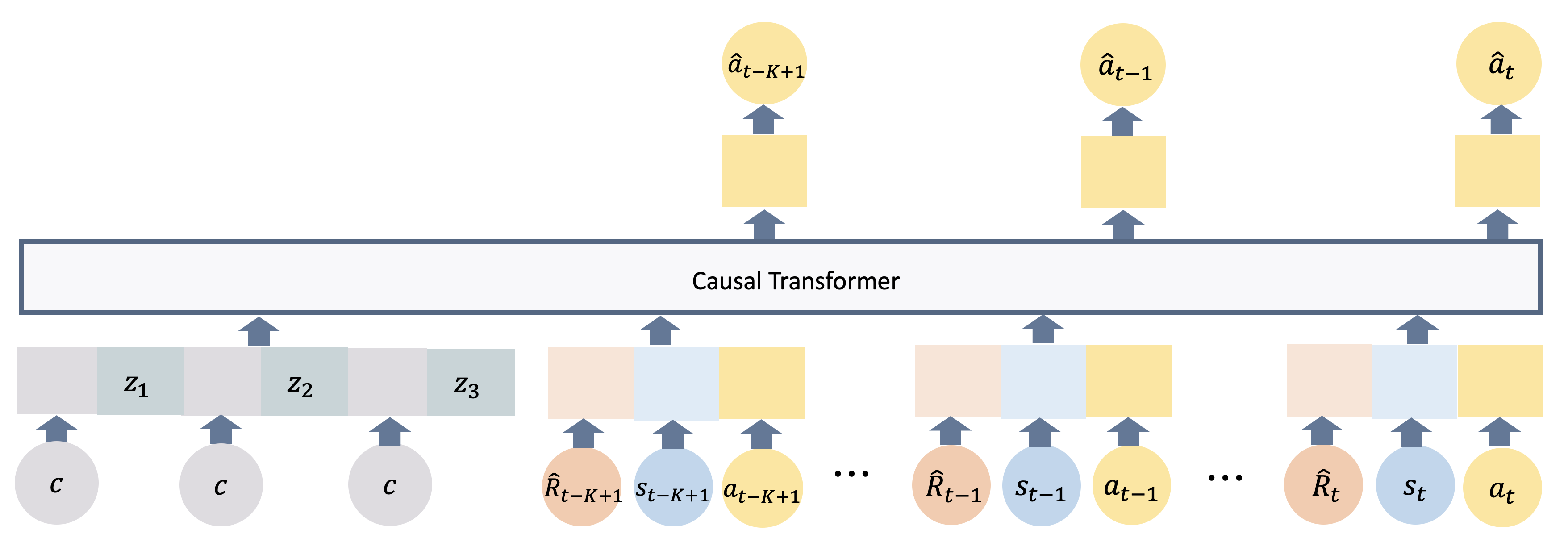}
        \caption{Architecture of the minimalist prompting decision transformer. At each time step $t$, the model receives the recent $K$ time step input trajectory prepended by a minimalist prompt (gray and green part) and outputs a sequence of actions until $\hat{\a}_t$. A minimalist prompt consists of a task parameter vector $[\mathbf{c}, \mathbf{c}, \mathbf{c}]$ and a learnable prompt $\mathbf{z}=[\mathbf{z_1}, \mathbf{z_2},\mathbf{z_3}]$.}
        \label{fig:model}
    \end{minipage}   
\end{figure}

To identify the essential factors of a demonstration prompt in generalization and further extract generalizable information from the supervision, we propose a minimalist prompting Decision Transformer (Fig. \ref{fig:model}). Given task parameter $\mathbf{c}$, a minimalist prompt consists of two parts: (1) A task parameter vector $[\mathbf{c}, \mathbf{c}, \mathbf{c}]$  (2) A learnable prompt $\mathbf{\z}=[\mathbf{\z_1}, \mathbf{\z_2}, \mathbf{\z_3}]$, where $\mathbf{\z_i}$ $(i=1,2,3)$ is a $n \times h$ dimensional vector corresponding to an embedding of $n$ tokens. The number of $\mathbf{\z_i}$ and $\mathbf{c}$ is designed to be three, paired with the three types of tokens $\widehat{R}, \s, \a$ in the input sequences.  

A minimalist prompting DT represents a policy $\pi_{\theta}(\a_t | \mathbf{w}_t, \mathbf{c}; \mathbf{z})$, where context $\mathbf{w}_t$ denotes the latest $K$ timestep input context at time step $t$.  
To train the policy on a task, the learnable prompt $\mathbf{z}$ is optimized together with the model weights $\mathbf{\theta}$ over the associated training set $\mathcal{D}=\{\tau^*_i\}_{i=1}^n$, thus the learning objective is
\begin{equation} \label{eq:task_prompt_dt_bc}
 \operatorname{argmin}_{\theta, z}\mathcal{L}(\mathbf{\theta}, \mathbf{z}; \mathcal{D}, \mathbf{c}) =\operatorname{argmin}_{\theta, z} \sum_{i=1}^{n} \mathcal{L} (\tau_{i}^{\star}; \pi_\theta(\cdot | \cdot, \mathbf{c}; \mathbf{z}))
\end{equation}

When training the policy $\pi_{\theta}(a_t | \mathbf{w}_t, \mathbf{c}; \mathbf{z})$ across $N$ tasks, the task parameter vector carries the task-specific information, whereas the learned prompt $\mathbf{z}$ captures the generic information shared by all training tasks. To see this, we can think of DT as a generator mapping latent variable $\mathbf{z}$ to actions, where $\mathbf{z}$ is adjusted to be broadly applicable to all of the training tasks. Therefore, although prompt $\mathbf{z}$ is {\bf not allowed} to be optimized during the test time, it is expected to be transferable to any unseen tasks from the same task distribution.

\subsection{Algorithms}
We now show how a minimalist prompting DT is trained and evaluated in contextual RL setting in Alg. \ref{alg:train} and \ref{alg:test}. Assume that we have $N$ training tasks $\mathcal{T}^{train}=\{T_i^{\text{train}}\}_{i=1}^N$, each of them has an associated offline trajectory dataset $\mathcal{D}^{\text{train}}_i=\{\tau_j\}_{j=1}^m$ and a task parameter $\mathbf{c}^{\text{train}}_i$. For each gradient step, the minimalist-prompting DT is optimized according to \eqref{eq:task_prompt_dt_bc} over a batch of trajectory segments and task parameters gathered from each training task.  

We evaluate a trained minimalist-prompting DT for each of $M$ test tasks $\mathcal{T}^{test}=\{T_i^{test}\}_{i=1}^M$ online. Since DT is inherently a return-conditioned policy, commanding the agent to perform a specific task should be conveyed as a desired target return $G_i$. To test the agent at its fullest capability, $G_i$ is usually set as the highest task return in task $i$, which therefore requires prior knowledge about the test task. In practice, for each test task, $G_i$ can be either set as the highest expert return or the highest possible return inferred from the upper bound of the reward function. With $G_i$ as its desired performance, the policy performs online evaluations by interacting with the test environment for $E$ episodes. The returns are then converted to the normalized score defined in \eqref{def:score} and averaged for across-task comparison.

\begin{algorithm}
\caption{Minimalist Prompting DT Training }\label{alg:train}
\hspace*{0 pt} \textbf{Input: }
$\theta$, $\mathbf{z}$ \Comment{Initialized parameters}\\
$\{\mathcal{D}^{\text{train}}_i, \mathbf{c}^{\text{train}}_i\}_{i=1}^N$ \Comment{Offline data and task parameters from $N$ training tasks}
\begin{algorithmic}
\For{each iteration}
\For{each gradient step}
    \State $\mathcal{B}= \emptyset$, $\mathcal{U}= \emptyset$
    \For{each task $T_i$} \Comment{Sample a batch of data}
        \State $\mathcal{B}_i = \{\tau_j\}_{j=1}^{|B_i|} \sim \mathcal{D}^{\text{train}}_i$ 
        \Comment{Sample a mini-batch of K-length trajectory segments 
        }
    \State $\mathcal{B} \leftarrow \mathcal{B} \cup \mathcal{B}_i$
    \State $\mathcal{U} \leftarrow \mathcal{U} \cup \{\mathbf{c}^{\text{train}}_i\}_{|B_i|}$
    \EndFor
    \State $[\mathbf{\theta}, \mathbf{z}] \leftarrow [\mathbf{\theta}, \mathbf{z}] -\alpha \nabla_{\mathbf{\theta}, z} \mathcal{L}\left(\mathbf{\theta}, \mathbf{z}; \mathcal{B},  \mathcal{U}\right)$ 
    \Comment {Update model weights and learnable prompt 
    }
    \EndFor
    
\EndFor
\end{algorithmic}
\hspace*{0 pt} \textbf{Output: }$\theta$, $\mathbf{z}$ \Comment{Optimized parameters}
\end{algorithm}

\begin{algorithm}
\caption{Minimalist Prompting DT Evaluation}\label{alg:test}
\hspace*{0 pt} \textbf{Input: } $\mathbf{\theta}, \mathbf{z}$ \Comment{Trained policy}\\
$\{\mathcal{M}_i=\left(\mathcal{S}_i, \mathcal{A}_i, \mathcal{P}_i, \rho_{0}^i, r_i, \gamma_i \right)\}_{i=1}^M$ 
\Comment{MDPs of $M$ test tasks}\\
$\{\mathbf{c}^{\text{test}}_i\}_{i=1}^M$ 
\Comment{Task parameters of $M$ test tasks}\\
$\{G^{\text{test}}_i\}_{i=1}^M$ 
\Comment{Target returns-to-go of $M$ test tasks}
\begin{algorithmic}
\For{each task $T_i$}
\State{$\mathbf{c} \leftarrow \mathbf{c}^{\text{test}}_i$}
\Comment{Get task parameter}
\State{$\widehat{R}_0 \leftarrow G^{\text{test}}_i$}
\Comment{Get task specific target returns-to-go}
\For{each test episode}
\State{$s_0 \sim \rho_i(s_0)$}
\Comment{Sample an initial state $s_0$}
\State{Initialize context $\mathbf{w}_t$ with $s_0$, $\widehat{R}_0$}
        \For{t=$0$, $T-1$}
        \State $\a_t \leftarrow \pi_{\theta}(\a_t | \mathbf{w}_t, \mathbf{c}; \mathbf{z})$ 
        \Comment{Get an action $\a_t$ using the policy}
        \State $\s_{t+1} \sim P_i(\s_{t+1}|\s_t, \a_t)$ 
        \Comment {Execute the action $\a_t$ and observe $\s_{t+1}$, $r_{t+1}$}
        \State $r_{t+1} = r_i(\s_t, \a_t)$
        \If{done}
            \State{break}
        \EndIf
        \Comment{Episode terminates}
        \State{$\widehat{R}_{t+1}=\widehat{R}_{t}-r_{t+1}$}
        \Comment{Update returns-to-go}
        \State{$\mathbf{w}_{t+1} \leftarrow \mathbf{w}_t \| (\a_t, \widehat{R}_{t+1}, \s_{t+1})$}
        \Comment{Update context}
        \EndFor
    \EndFor
\EndFor
\end{algorithmic}
\end{algorithm}

\section{Experiments and Results}
In this section, we evaluate the zero-shot generalization ability of the minimalist prompting DT over a set of benchmark control and manipulation tasks and compare it with the demonstration-prompting counterpart. We start by introducing the experiment setting and then proceed to the result analysis.

\subsection{Environments and tasks}
The experiments are conducted on two widely used benchmarks: control benchmark MACAW \cite{macaw} for offline meta-reinforcement learning and manipulation benchmark Meta-world \cite{meta-world} for meta-reinforcement learning. Specifically, we evaluate our approach on the following five problems: 

{\bf Cheetah-Velocity} A simulated half-cheetah learns to run at varying goal velocities. The tasks differ in the reward functions $r_g(s,a)$ which is parameterized by the goal velocity $g$. We train in 35 training tasks and evaluate in 5 test tasks with unseen goal velocities. The half-cheetah has 6 joints and 20-dimensional state space which includes the position and velocity of the joints, the torso pose, and COM (Center of Mass) of the torso. Note that the goal velocity is not observable to the agent.

{\bf Ant-Direction} A simulated ant learns to run in varying 2D directions. The tasks differ in the reward functions $r_g(s,a)$ which is parameterized by the goal direction $g$. We train in 45 training tasks and evaluate in 5 test tasks with unseen goal directions. The ant has 8 joints and 27-dimensional state space which is represented by the position and velocity of the joints and pose of the torso. Note that the goal direction is not observable to the agent.

By following the same data preparation methods in \cite{macaw} and \cite{prompt-dt}, the offline data for each task of Cheetah-Velocity and Ant-Direction are drawn from the lifelong replay buffer of a Soft Actor-Critic \cite{sac} agent trained on it, where the first, middle, and last 1000 trajectories correspond to the random, medium, and expert data respectively.

{\bf ML1-Reach}, {\bf ML1-Push}, {\bf ML1-Pick-Place} A Sawyer robot learns to reach, push, and pick and place an object at various target positions. The tasks within each problem vary in the reward functions $r_g(s,a)$ and initial state distributions $\rho_o(s_0)$, where $r_g(s,a)$ is parameterized by the target position $g$ and $\rho_o(s_0)$ is parameterized by the initial object position $o$. The state space is 39 dimensional including the pose of the end-effector and the pose of the object in the recent two time steps. Note that the target position is not part of the state. For each problem, there are 45 training tasks and 5 test tasks with unseen target positions. The expert demonstration trajectories for each task are generated by running the provided scripted expert policy. 

\subsection{Baselines}
To understand which factors contribute to generalization, we compare the full minimalist-prompting DT (Task-Learned-DT) with four baseline approaches, including three variants of minimalist-prompting DT, one state-of-art few-shot prompt DT method, and the original decision transformer:

{\bf Minimalist-prompting DT with both task parameter prompt and learned prompt (Task-Learned-DT)} This is our proposed method in its fullest form. In the experiments, we use goal velocity, goal direction, and target position as the task parameter for Cheetah-Velocity, Ant-Direction, and ML1 tasks respectively. 

{\bf Minimalist-prompting DT with task parameter prompt only (Task-DT)} To study the effect of the learned prompt on zero-shot generalization, we remove it from Task-Learned-DT and only keep the task parameter prompt. This can be thought of as a special case of Task-Learned-DT where the learned token has length 0.

{\bf Pure-Learned-DT} To study the effect of the task parameter prompt in zero-shot generalization, we remove it from Task-Learned-DT and only keep the learned prompt.

{\bf Trajectory-prompting DT (Trajectory-DT)} \cite{prompt-dt} is a state-of-the-art few-shot policy learning method that shares a prompt DT architecture similar to ours. However, this method requires segments of expert demonstration trajectories as prompt in the test environments, which are unrealistic in most real-world applications.

{\bf Decision transformer (DT)} We apply the original decision transformer \cite{dt} to our experiment setting. By excluding any prompts, we train the original decision transformer in multiple tasks and test it in other unseen tasks. This helps us understand the role of prompts in generalization.  

\subsection{Experimental results}
We investigate the zero-shot generalization ability of Task-Learned-DT and identify the essential factors by comparing it to the baselines. The zero-shot generalization performance is evaluated online under the mean normalized score in unseen test tasks. If not mentioned explicitly, all methods are trained on the same expert dataset in each task. In particular, Trajectory-DT is conditioned on the expert trajectory prompts sampled from the corresponding expert dataset. For a fair comparison, all algorithms are run across the same three seeds. Our experiments aim to empirically answer the following questions:

\subsubsection{How does minimalist-prompting DT generalize compare to its demonstration-prompting counterpart?}
\begin{figure}[t!]
        \centering 
        \includegraphics[width=\columnwidth]{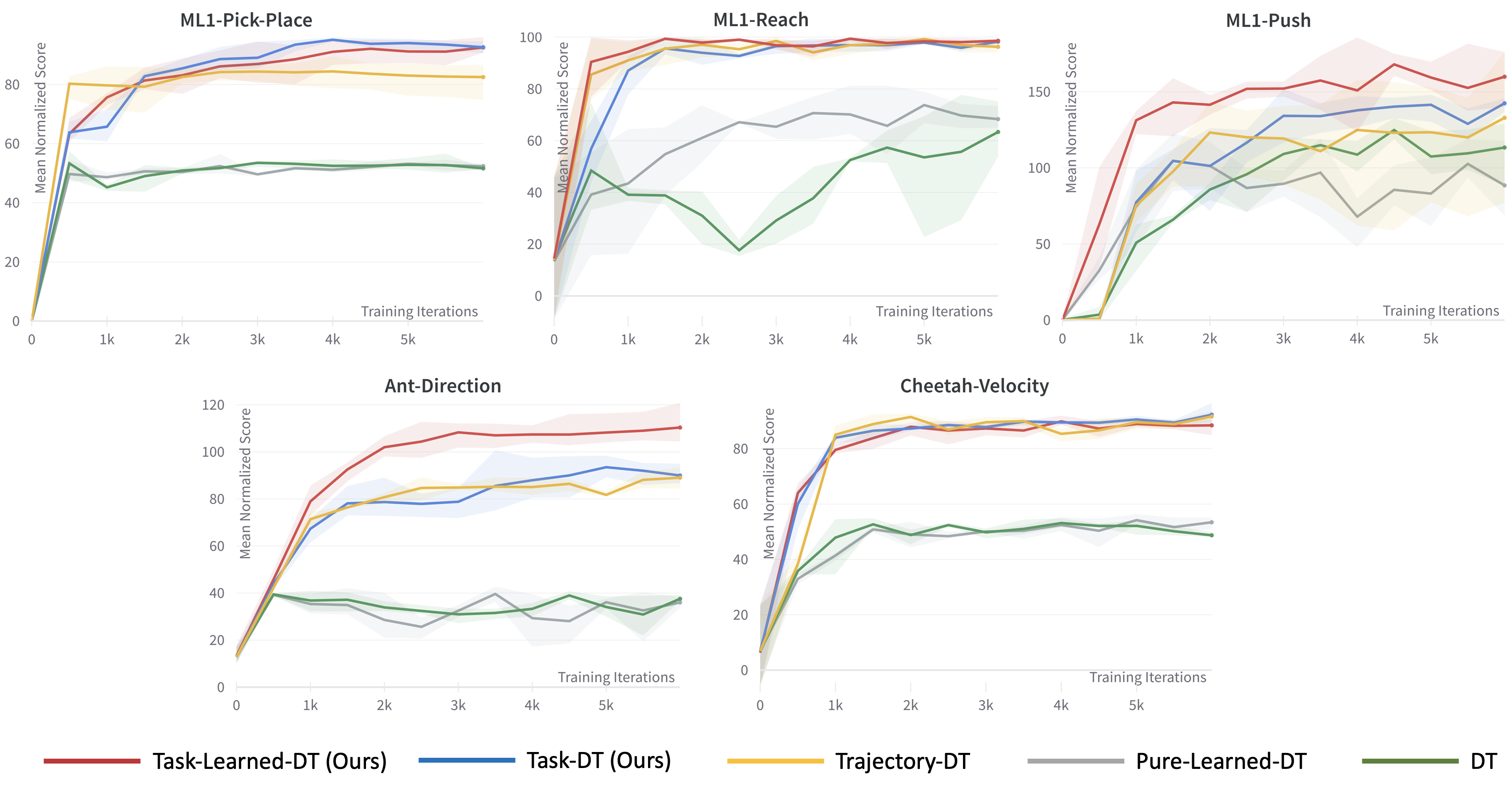}
        \caption{Comparing full minimalist-prompting DT (Task-Learned-DT) with four baselines: Task-DT, Trajectory-DT, Pure-Learned-DT and DT. The zero-shot performance of each algorithm is evaluated through the entire training process on five benchmark problems and reported under the mean normalized score. Shaded regions show one standard deviation of three seeds. }
    \label{fig:main_results}
\end{figure}

We compare Task-Learned-DT with the baselines on all five benchmark problems. The results are shown in Figure \ref{fig:main_results} and Table \ref{main-table}, in which Task-Learned-DT and Task-DT consistently outperform Trajectory-DT across all five problems and exceed it by a large margin on ML1-Pick-Place, ML1-Push, and Ant-Direction (Table \ref{performance-improve-table}). This indicates that the task parameters have precisely encoded sufficient task variations to specify the new task. Trajectory-DT tries to extract the same information from the prompting expert trajectories but performs worse than directly feeding it. In other words, prompting the problem solutions to the agent is inferior to prompting the task parameters.

To better understand how generalization happens via the proposed prompt, we further investigate the role of the learnable prompt. By comparing Task-Learned-DT and Task-DT, we find that the learned prompt improves generalization in most tasks, yet on ML1-Pick-Place and Cheetah-Velocity, adding a learned prompt slightly hurts the generalization (Table \ref{performance-improve-table}). However, we observed that the learned prompt will improve the generalization when tuned to an appropriate length. We thereby hypothesize that to benefit generalization, the length of the learned prompt should capture the dimension of the common representation of the tasks. In other words, one should choose a longer learned prompt if the solutions to different tasks share more similarities. We will look into this later via ablating the length of the learned prompt in Table \ref{prompt-length-table}. 

Next, we study the generalization ability provided by the task parameters by comparing Task-Learned-DT with Pure-Learned-DT and DT. The results show that Pure-Learned-DT performs similarly to DT and is significantly worse than Task-Learned-DT. In other words, learning a prompt without task parameters is almost equivalent to having no prompt at all. This implies that without the guidance of task parameter prompts, extracting the representation of a shared skill alone is not sufficient to perform well in a new task. In other words, the task-specific prompt is necessary for our method to outperform Trajectory-DT, although not always contribute to a larger performance gain.

\begin{table}[t!]
  \caption{Comparison on five benchmark problems}
  \label{main-table}
  \centering
  \resizebox{\columnwidth}{!}{
  \begin{tabular}{l | l l l l l}
    \toprule
    Problem     & Task-Learned-DT (Ours)  & Task-DT (Ours) & Trajectory-DT & Pure-Learned-DT & DT \\
    \midrule
    ML1-Pick-Place & 92.52 $\pm$ 3.03  & {\bf 92.58 $\pm$ 1.99}  & 82.50 $\pm$ 6.73 & 52.39 $\pm$ 1.21  &  51.63 $\pm$ 0.22\\
    ML1-Reach & {\bf 98.58 $\pm$ 1.02}  & 98.25 $\pm$ 1.04  & 96.24 $\pm$ 2.42 & 68.33 $\pm$ 4.49 & 63.35 $\pm$ 10.86 \\
    ML1-Push & {\bf 159.88 $\pm$ 18.23}  & 142.42 $\pm$ 4.89 & 132.91 $\pm$ 50.58 & 88.50 $\pm$ 25.90  & 113.39 $\pm$ 21.96 \\
    Ant-Direction  & {\bf 110.35 $\pm$ 9.10} & 89.96 $\pm$ 4.41  & 89.04 $\pm$ 4.43 &  36.00 $\pm$ 2.28 & 37.55 $\pm$ 1.36 \\
    Cheetah-Velocity & 88.43 $\pm$ 3.15 & {\bf 92.29 $\pm$ 3.81} & 91.64 $\pm$ 1.83 & 53.42 $\pm$ 2.17  & 48.72 $\pm$ 1.02 \\
    \bottomrule
  \end{tabular}}
\end{table}

\begin{table}[t!]
  \caption{Performance improvement on five benchmark problems}
  \label{performance-improve-table}
  \centering
  \resizebox{0.9\columnwidth}{!}{
  \begin{tabular}{l | l | l}
    \toprule
    Problem  & $\max$(Task-DT,Task-Learned-DT) - Trajectory-DT & Task-Learned-DT - Task-DT\\
    \midrule
    ML1-Pick-Place & 10.08 & -0.06\\
    ML1-Reach & 2.34 & 0.33\\
    ML1-Push &  26.97 & 17.46\\
    Ant-Direction  & 21.31 & 20.39\\
    Cheetah-Velocity & 0.65 & -3.86\\
    \bottomrule
  \end{tabular}}
\end{table}

\subsubsection{How does the length of the learned prompt affect minimalist-prompting DT?}
We vary the token number of the learned prompt in the range of $\{0,3,15,30\}$ on ML1-Pick-Place, ML1-Push, and Ant-Direction to study its impact on generalization (Table \ref{prompt-length-table}).  The learned prompt affects the performance to varying degrees depending on the problem domain.  We observed that having a longer learned prompt does not always lead to better performance. A learned prompt of length 15 or 30 often achieves satisfactory performance even if not necessarily the best. 

\begin{table}[t!]
  \caption{Ablation on learned prompt length}
  \label{prompt-length-table}
  \centering
  \begin{tabular}{l | l l l l }
    \toprule
    Problem   & 0 & 3 & 15 & 30 \\
    \midrule
    ML1-Pick-Place & {\bf 92.58 $\pm$ 1.99} & 91.31 $\pm$ 5.36 & 92.52 $\pm$ 3.03  & 82.66 $\pm$ 8.62 \\
    ML1-Push & 142.42 $\pm$ 4.89  &  134.46 $\pm$ 14.46 & 140.16 $\pm$ 10.85 & {\bf 159.88 $\pm$ 18.23} \\
    Ant-Direction &  89.96 $\pm$ 4.41 & 98.28 $\pm$ 8.30 & {\bf 110.35 $\pm$ 9.10} & 100.04 $\pm$ 5.26 \\
    \bottomrule
  \end{tabular}
\end{table}

\subsubsection{Can minimalist-prompting DT master multiple skills as well as performing zero-shot generalization?} 

To better understand the role of task parameters and the learned prompt, we consider a more challenging setting where we train the agent to master distinct skills and evaluate its zero-shot generalization to new tasks. Meta-world ML10 benchmark \cite{meta-world} provides 10 meta-task for training. Each of them corresponds to a distinct skill such as closing a drawer and pressing a button. Varying the object and goal positions within each meta-task yields 50 tasks. We then randomly draw 5 held-out tasks for testing in each meta-task and train on the others. We use the same task parameter as ML1. 

Table \ref{ml10-table} compares zero-shot generalization performance of Task-Learned-DT, Task-DT, and Trajectory-DT in this setting. Among them, Task-Learned-DT outperforms Trajectory-DT by a large margin. This indicates that minimalist-prompting DT can act as a generalist agent with zero-shot generalization capability. Although the task parameters of each meta-task has a similar form, minimalist-prompting DT is able to interpret them according to the context environment and modulate itself to generalize to the corresponding new task. In addition, the performance gain of Task-Learned-DT over Task-DT indicates that the learned prompt becomes more necessary in this multi-skill generalization setting. We interpret this as mastering more skills requires grasping their commonalities better.

\begin{table}[t!]
  \caption{Comparison on ML10 tasks}
  \label{ml10-table}
  \centering
  \begin{tabular}{l | l l l}
    \toprule
    Problem     & Task-Learned-DT (Ours)  & Task-DT (Ours) & Trajectory-DT \\
    \midrule
    ML10 & {\bf 74.56 $\pm$ 4.11}  &  57.85 $\pm$ 0.91  & 63.14 $\pm$ 5.65 \\
    \bottomrule
  \end{tabular}
\end{table}

\section{Limitations}
As shown in Table \ref{prompt-length-table}, the length of the learned prompt is a problem-dependent hyperparameter for our method to reach its best performance. Our approach also inherits limitations from Decision Transformer: (1) Since DT is optimizing a behavior cloning (BC) objective, it may suffer from a performance drop as the training trajectories become sub-optimal or even random \cite{offline_bc} (Table \ref{quality-table}). (2) Deploying a DT-based model to a new task requires setting up a target returns-to-go in advance to reflect the desired highest performance, which needs some prior knowledge about the task.

\section{Conclusion}
In this work, we study the problem of what are the essential factors in the demonstration prompts for generalization. By identifying them, we hope to get rid of the assumption that demonstrations must be accessible during deployment, which is unrealistic in most real-world robotics applications. Under the framework of contextual RL, we empirically identify the task parameters as the necessary information for a decision transformer to generalize to unseen tasks. Built on this observation, we additionally introduce a learnable prompt to further boost the agent's generalization ability by extracting the shared generalizable information from the training supervision. We demonstrated that the proposed model performs better than its demonstration prompting counterpart across a variety of control, manipulation, and planning benchmark tasks.

\section{Appendix}
\input{appendix.tex}

\bibliographystyle{unsrtnat}
\bibliography{references}

\end{document}

%% file: appendix.tex
\newcounter{alphasect}
\def\alphainsection{0}

\let\oldsection=\section
\def\section{%
  \ifnum\alphainsection=1%
    \addtocounter{alphasect}{1}
  \fi%
\oldsection}%

\renewcommand\thesection{%
  \ifnum\alphainsection=1%
    \Alph{alphasect}
  \else%
    \arabic{section}
  \fi%
}%

\newenvironment{alphasection}{%
  \ifnum\alphainsection=1%
    \errhelp={Let other blocks end at the beginning of the next block.}
    \errmessage{Nested Alpha section not allowed}
  \fi%
  \setcounter{alphasect}{0}
  \def\alphainsection{1}
}{%
  \setcounter{alphasect}{0}
  \def\alphainsection{0}
}%

\begin{alphasection}

\section{Additional Experiments and Ablations}
\subsection{Additional experiments on D4RL Maze2D}
In addition to control and manipulation tasks, we investigate whether minimalist-prompting DT can also perform well in navigation tasks. We experiment with Maze2D environment from the offline RL benchmark D4RL \cite{d4rl}. The Maze2D domain is a continuous 2D navigation task that requires a point mass to navigate to a goal location. We use the medium maze layout and create 21 training tasks and 5 test tasks. The tasks differ in the goal locations while the initial locations in each task are selected randomly for both training and evaluation episodes. The goal locations of the test tasks are unseen during the training. The agent has a 4-dimensional state space which includes its current 2D position and 2D velocity. The reward function $r_g (s, a)$ of each task is parameterized by the corresponding goal location $g$. Note that the goal location is not observable to the agent.

We only use the goal location as the task parameter but not the initial location which is akin to common real-world scenarios where a human is asking a robot to navigate to a goal location and the robot is expected to know its starting location without explicit instructions. This setup poses more challenging generalization requirements during evaluation where the agent should not only generalize to unseen goal locations but also unseen initial locations.

We compare Task-Learned-DT with four baselines in terms of zero-shot generalization in Table \ref{table:maze2d}. It is observed that Task-Learned-DT still outperforms Trajectory-DT by a large margin. All methods are trained on the expert demonstration trajectories generated by running the provided expert controller \cite{d4rl}.

\subsection{Comparison on medium and random data} 
To study the zero-shot generalization performance of Task-Learned-DT when expert demonstrations are not available, we train Task-Learned-DT, Task-DT, and Trajectory-DT on the medium and random trajectories on Cheetah-Velocity (Table \ref{quality-table}). Trajectory-DT is conditioned on the trajectory segments of the same quality in both training and test tasks. Although all of these methods experience a significant performance drop, our methods still perform better than Trajectory-DT. In particular, Task-Learned-DT outperforms Task-DT on medium and random data. This suggests that the learned prompt provides some robustness against the noises in the data. Based on the theoretical analysis of prior work \cite{offline_bc}, the performance drop of DT-based methods on sub-optimal data is possibly attributed to the limitations of behavior cloning.

\subsection{Does minimalist-prompting DT perform well in sparse reward settings?} 
To investigate whether minimalist-prompting DT relies on the densely populated rewards, we create a delayed return version of ML1-Pick-Place, Ant-Direction,  and Cheetah-Velocity following the original DT paper \cite{dt}. The zero-shot generalization performance is evaluated in Table \ref{sparse-reward-table}. We observed that both Task-Learned-DT and Task-DT are minimally affected by the sparsity of the rewards in most cases.

\begin{table}[t!]
  \caption{Comparison on Maze2D}
  \label{table:maze2d}
  \centering
  \resizebox{\columnwidth}{!}{
  \begin{tabular}{l | l l l l l}
    \toprule
    Problem     & Task-Learned-DT (Ours)  & Task-DT (Ours) & Trajectory-DT & Pure-Learned-DT & DT \\
    \midrule
    Maze2D & {\bf 23.00 $\pm$ 2.56}  &  17.58 $\pm$ 2.49  & 18.34  $\pm$ 2.89 &  2.04 $\pm$ 0.37 &  2.21 $\pm$ 1.30 \\
    \bottomrule
  \end{tabular}}
\end{table}

\begin{table}[t!]
  \caption{Ablating different data qualities on Cheetah-Velocity}
  \label{quality-table}
  \centering
  \begin{tabular}{l | l l l}
    \toprule
    Quality     & Task-Learned-DT (Ours)  & Task-DT (Ours) & Trajectory-DT \\
    \midrule
    Expert & 88.43 $\pm$ 3.15 & {\bf 92.29 $\pm$ 3.81} & 91.64 $\pm$ 1.83 \\
    Medium & {\bf 20.04 $\pm$ 0.16} & 19.30 $\pm$ 1.74 & 16.58 $\pm$ 3.72 \\
    Random & {\bf 21.95 $\pm$ 4.22} & 17.41 $\pm$ 8.46 & 19.86 $\pm$ 3.47 \\
    \bottomrule
  \end{tabular}
\end{table}

\begin{table}[t!]
  \caption{Ablation on sparse rewards}
  \label{sparse-reward-table}
  \centering
  \resizebox{0.9\columnwidth}{!}{
  \begin{tabular}{l | l l | l l}
    \toprule
    Problem  & \multicolumn{2}{c|}{Task-Learned-DT (Ours)} & \multicolumn{2}{c}{Task-DT (Ours)}   \\
    & Original (Dense) & Delayed (Sparse) &  Original (Dense) & Delayed (Sparse)\\
    \midrule
    ML1-Pick-Place & 92.52 $\pm$ 3.03 & 93.58 $\pm$ 3.24 & 92.58 $\pm$ 1.99 & 92.20 $\pm$ 3.44 \\
    Ant-Direction  & 110.35 $\pm$ 9.10 & 102.15 $\pm$ 1.92 & 89.96 $\pm$ 4.41  & 88.92 $\pm$ 2.09 \\
    Cheetah-Velocity & 88.43 $\pm$ 3.15 & 89.63 $\pm$ 1.28  & 92.29 $\pm$ 3.81 & 91.41 $\pm$ 3.70\\
    \bottomrule
  \end{tabular}}
\end{table}

\subsection{How does minimalist-prompting DT perform on the seen tasks?}
We evaluate Task-Learned-DT and four baselines on the training tasks of five benchmark problems and report the performance in terms of mean normalized score in Table \ref{table:id-main} and Figure \ref{fig:main_train_results}. In ML1-Reach, ML1-Push, and Cheetah-Velocity, Task-Learned-DT (Task-DT) outperforms Trajectory-DT and maintains high performance in the rest two tasks. This observation shows that Task-Learned-DT (Task-DT) is able to improve zero-shot generalization without harming performance much on the seen tasks.

\begin{table}[t!]
  \caption{Seen task performance on five benchmark problems}
  \label{table:id-main}
  \centering
  \resizebox{\columnwidth}{!}{
  \begin{tabular}{l | l l l l l}
    \toprule
    Problem     & Task-Learned-DT (Ours)  & Task-DT (Ours) & Trajectory-DT & Pure-Learned-DT & DT \\
    \midrule
    ML1-Pick-Place & 98.41 $\pm$ 2.36  & 99.63 $\pm$ 0.32  & {\bf 100.28 $\pm$ 0.08} & 71.27 $\pm$ 1.09  &  76.94 $\pm$ 1.02\\
    ML1-Reach & 99.76 $\pm$ 0.23  & {\bf 99.81 $\pm$ 0.06}  & 98.24 $\pm$ 2.68 & 72.13 $\pm$ 5.87 & 62.69 $\pm$ 3.92 \\
    ML1-Push & {\bf 142.97 $\pm$ 6.77}  & 133.30 $\pm$ 7.41 & 136.65 $\pm$ 3.06 & 99.57 $\pm$ 6.35  & 101.94 $\pm$ 3.48 \\
    Ant-Direction  & 89.57 $\pm$ 2.48 & 84.96 $\pm$ 2.05  & {\bf 92.76 $\pm$ 0.10} &  26.36 $\pm$ 1.48 & 29.53 $\pm$ 1.21 \\
    Cheetah-Velocity & 96.88 $\pm$ 0.60 & {\bf 97.53 $\pm$ 0.53} & 96.41 $\pm$ 1.21 & 48.70 $\pm$ 1.10  & 47.19 $\pm$ 3.68 \\
    \bottomrule
  \end{tabular}}
\end{table}

\begin{figure}[t!]
        \centering 
        \includegraphics[width=\columnwidth]{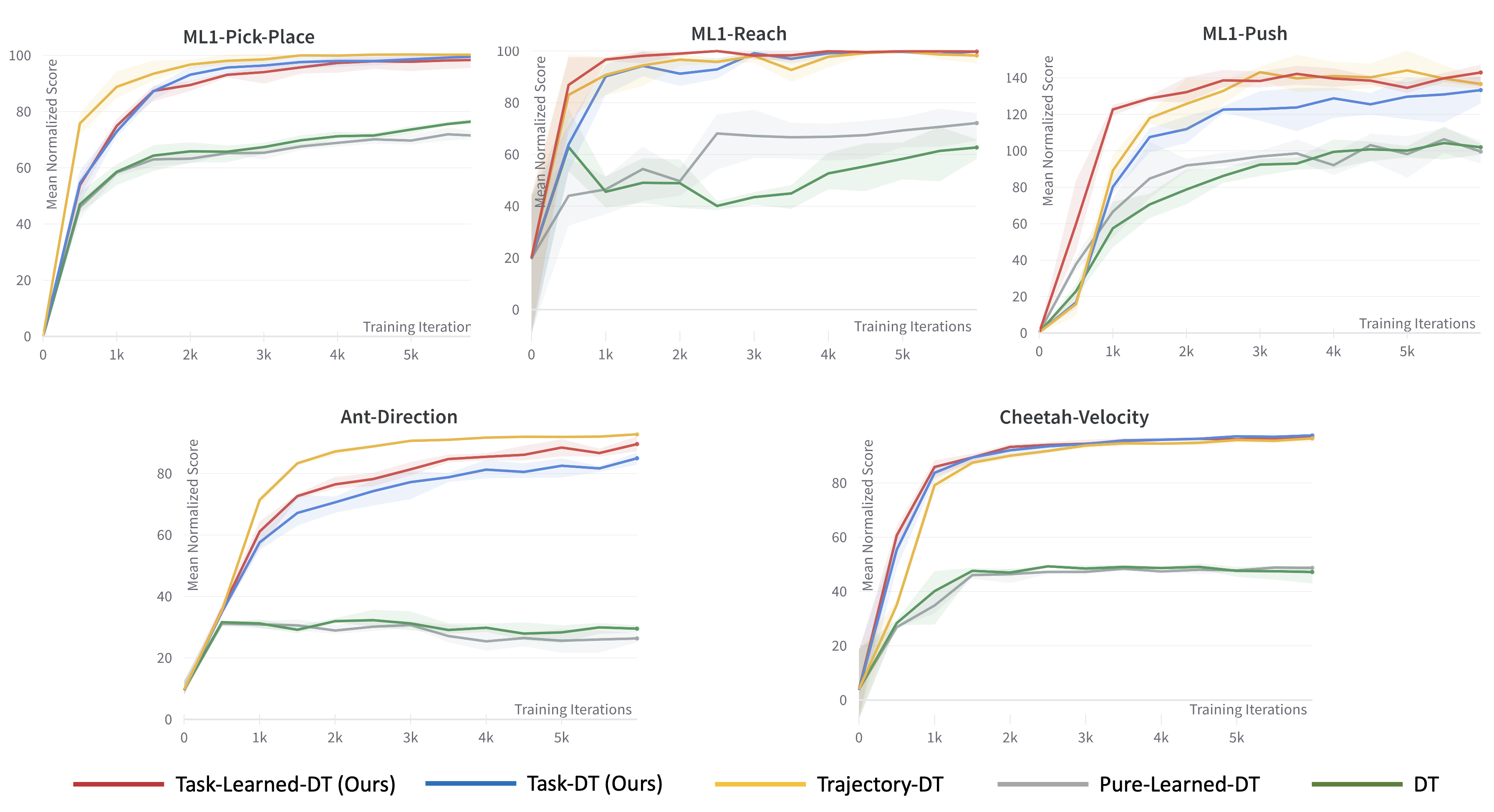}
        \caption{Comparing full minimalist-prompting DT (Task-Learned-DT) with four baselines Task-DT, Trajectory-DT, Pure-Learned-DT and DT on training tasks. The seen task performance of each algorithm is evaluated through the entire training process on five benchmark problems and reported under the mean normalized score. Each training task is evaluated for 20 episodes. Shaded regions show one standard deviation of three seeds. }
    \label{fig:main_train_results}
\end{figure}

 \section{Task and Dataset Details
}
\subsection{Task parameters}
To help understand our approach, we provide concrete examples of the task parameters we adopted in each problem in Table \ref{table:task_parameter}. The task parameters we experiment with capture the minimally sufficient task variations. 

\begin{table}[t!]
  \caption{Examples of task parameters}
  \label{table:task_parameter}
  \centering
    \begin{tabular}{l|l l}
    \toprule
    Problem & \multicolumn{2}{c}{Task Parameter}\\
    & Problem-specific Meaning &  Example \\
    \hline
    Cheetah-Velocity & goal velocity & [0.08] \\
    Ant-Direction & goal direction & [1.20] \\
    ML1-Pick-Place & target position &  [0.01, 0.84, 0.22] \\
    ML1-Reach & target position & [-0.07,  0.81,  0.08] \\
    ML1-Push & target position & [0.02, 0.84, 0.02] \\
    ML10  & target position & [-0.30,  0.53,  0.13] \\ 
    Maze2D & goal location & [1,1] \\
\bottomrule
    \end{tabular}
\end{table}

\subsection{Task splits}
We summarize the splits of training and test tasks for each problem in Table \ref{table:task-splits}. For Cheetah-Velocity and Ant-Direction, we follow the splits provided by \cite{prompt-dt}. For the rest problems, we randomly sample a subset of tasks for evaluation. 

In particular, we list the goal locations for training and test tasks in Maze2D. Maze2D with the medium layout is a $6 \times 6$ maze. We choose goal locations as the integer coordinates of all 26 empty cells in the maze and split them into training and test sets as follows:

The list of 21 training tasks:
\begin{itemize}
    \item Task indices:
    0, 2, 3, 4, 5, 7, 8, 9, 11, 13, 14, 15, 16, 17, 18, 19, 20, 21, 22, 23, 24
    \item Goal locations: 
    [1,1], [1,5], [1,6], [2,1], [2,2], [2,5],
    [2,6], [3,2], [3,4], [4,2], [4,4], [4,5], [4,6], [5,1], [5,3], [5,4], [5,6], [6,1],
    [6,2], [6,3], [6,5]
\end{itemize}
The list of 5 test tasks:
\begin{itemize}
    \item Task indices:
    1, 6, 10, 12, 25 
    \item Goal locations: 
    [1,2], [2,4], [3,3], [4,1], [6,6]
\end{itemize}

\begin{table}[t!]
  \caption{Task splits}
  \label{table:task-splits}
  \centering
  \begin{tabular}{l | l | l}
  \toprule
  Problem & Number of training tasks &  Number of test tasks \\
  \midrule
  Cheetah-Velocity & 35 & 5 \\
  Ant-Direction & 45 & 5 \\
  ML1-Pick-Place & 45 & 5 \\
  ML1-Reach & 45 & 5 \\
  ML1-Push & 45 & 5 \\
  ML10 & 450 (45 for each meta-task) & 50 (5 for each meta-task) \\
  Maze2D & 21 & 5 \\
  \bottomrule
  \end{tabular}
\end{table}

\subsection{Data generation}
For Cheetah-Velocity and Ant-Direction, we use the expert data released by \cite{prompt-dt} for training and evaluation. The medium and random data are generated from the replay buffer transitions provided by \cite{macaw} and follow the same instructions. For ML1 and ML10, the expert trajectories for each task are generated by running the expert policy provided by Meta-World \cite{meta-world}. For Maze2D, the expert trajectories for each task are generated by running the expert controller provided by D4RL \cite{d4rl}.

We collect a set of trajectories for each training and test task and randomly sample a subset for Trajectory-DT to extract trajectory prompts. Note that the trajectories for the test tasks are only used by Trajectory-DT during the evaluation. We list the number of demonstration trajectories collected per task for each problem in Table \ref{table:traj-num}.

\begin{table}[t!]
  \caption{Number of trajectories per task}
  \label{table:traj-num}
  \centering
  \begin{tabular}{l | l | l}
  \toprule
  Problem & Number of trajectories per task & Number of prompt trajectories per task\\
  \midrule
  Cheetah-Velocity & 999 & 5\\
  Ant-Direction & 1000 & 5\\
  ML1-Pick-Place & 100 & 5\\
  ML1-Reach & 100 & 5\\
  ML1-Push & 100 & 5\\
  ML10 & 100 & 5\\
  Maze2D & 100 & 5\\
  \bottomrule
  \end{tabular}
\end{table}

\subsection{ML10 meta-tasks}
To show the diversity of the skills we train the agent to master in ML10, we include a description of each of the 10 meta-tasks in Table \ref{table:ml10-meta-tasks}. Please refer to \cite{meta-world} for more details.

\begin{table}[t!]
  \caption{ML10 Meta-tasks}
  \label{table:ml10-meta-tasks}
  \centering
  \resizebox{\columnwidth}{!}{
  \begin{tabular}{l | l}
  \toprule
  Meta-task & Description \\
  \midrule
    reach-v2 & Reach a goal position. Randomize the goal positions. \\
    push-v2 & Push the puck to a goal. Randomize puck and goal positions. \\
    pick-place-v2 & Pick and place a puck to a goal. Randomize puck and goal positions. \\
    door-open-v2 & Open a door with a revolving joint. Randomize door positions. \\
    drawer-close-v2 & Push and close a drawer. Randomize the drawer positions. \\
    button-press-topdown-v2 & Press a button from the top. Randomize button positions. \\
    peg-insert-side-v2 & Insert a peg sideways. Randomize peg and goal positions. \\
    window-open-v2 & Push and open a window. Randomize window positions. \\
    sweep-v2 & Sweep a puck off the table. Randomize puck positions. \\
    basketball-v2 &  Dunk the basketball into the basket. Randomize basketball and basket positions. \\
    \bottomrule
  \end{tabular}}
\end{table}

 \section{Hyperparameters and Implementation Details}

\subsection{Hyperparameters}
We describe the hyperparameters shared by Task-Learned-DT, Task-DT, Trajectory-DT, Pure-Learned-DT, and DT in Table \ref{table: common_hyper}. For a fair comparison, we set most of the hyperparameter values to be the same as \cite{prompt-dt}. The evaluation results are reported when the maximum training iterations are reached and the training converged. All of the methods are run across the same three random seeds 1,6,8.

We show the hyperparameters specific to our method and Trajectory-DT in Table \ref{table: our_hyper} and Table \ref{table: traj_hyper} respectively. Trajectory-DT assembles trajectory segments of length $H$ from $J$ episodes as a trajectory prompt. We set the values of $H$ and $J$ as the ones reported in \cite{prompt-dt} in Cheetah-Velocity, Ant-Direction, and ML1-Reach. The rest problems use the same values as ML1-Reach.

\begin{table}[t!]
  \caption{Common hyperparameters}
  \label{table: common_hyper}
  \centering
    \begin{tabular}{l|l}
    \toprule
    Hyperparameter & Value \\
    \hline
    Context length K  & 20 \\
    Number of layers & 3 \\
    Number of attention heads & 1 \\
    Embedding dimension & 128 \\
    Nonlinearity & ReLU \\
    Dropout & 0.1 \\
    Learning rate & $10^{-3}$ when learned prompt length = 30 \\
    & $10^{-4}$ otherwise\\
    Weight decay  & $10^{-4}$ \\
    Warmup steps & $10^4$ \\
    Training batch size for each task & 16 \\
    Number of evaluation episodes for each task & 20 \\
    Number of gradient steps per iteration & 10 \\
    Maximum training iterations & 15500 ML10 \\
    & 6000 otherwise \\
    Random seeds & 1,6,8 \\
    \bottomrule
    \end{tabular}
\end{table}

\begin{table}[t!]
  \caption{Problem-specific hyperparameters of Task-Learned-DT and Pure-Learned-DT}
  \label{table: our_hyper}
  \centering
    \begin{tabular}{l|l}
    \toprule
    Problem & Learned Prompt Length \\
    \hline
    Cheetah-Velocity & 15 \\
    Ant-Direction & 15 \\
    ML1-Pick-Place & 15 \\
    ML1-Reach & 30 \\
    ML1-Push & 30 \\
    ML10 & 30 \\
    Maze2D & 30 \\
    \bottomrule
    \end{tabular}
\end{table}

\begin{table}[t!]
  \caption{Problem-specific hyperparameters of Trajectory-DT}
  \label{table: traj_hyper}
  \centering
    \begin{tabular}{l|l|l}
    \toprule
    Problem & The Number of Episodes $J$ & Trajectory Segment Length $H$ \\
    \hline
    Cheetah-Velocity & 1 & 5 \\
    Ant-Direction & 1 & 5 \\
    ML1-Pick-Place & 1 & 2 \\
    ML1-Reach & 1 & 2 \\
    ML1-Push & 1 & 2 \\
    ML10  & 1 & 2 \\
    Maze2D & 1 & 2 \\
    \bottomrule
    \end{tabular}
\end{table}

\subsection{Target returns-to-go}
In the experiments, we set the target returns-to-go for each task as the highest return of the associated expert demonstration trajectories. This is possible because we train with expert trajectories and Trajectory-DT requires expert trajectories even in the test tasks. However, setting the target returns-to-go according to the expert performance is not necessary. We can also set it based on our prior knowledge about the task, e.g. set it as the highest possible
return inferred from the upper bound of the reward function, or set it as the return representing a satisfactory performance.
\subsection{Implementation}
We implement our method based on the released repository of \cite{prompt-dt}. 

\subsection{Training}
Our model is trained using NVIDIA A100 Tensor Core GPUs. Training our model using a single A100 GPU on a benchmark problem takes about 3 hours for typically 6000 training iterations. Evaluating the model in training and test tasks every 500 iterations throughout the training process takes another 3 hours due to the online interactions with the simulated environments.   

\section{Failure modes}
By visualizing the trained policy performing the test tasks, we have observed that most failures occur when the agent has difficulty in precisely achieving unseen goal locations or directions, rather than struggling with mastering basic skills such as running or holding the block. This implies that including the task variations in the prompt or having the capability to extract task variations from the prompt is important to guarantee the task's success. 

\end{alphasection}